\documentclass[letterpaper, 10pt, conference]{latex_template/ieeeconf}  

\IEEEoverridecommandlockouts                              

\usepackage{graphics} 
\usepackage{epsfig} 
\usepackage{amsmath} 
\usepackage{amssymb}  

\usepackage{booktabs}
\usepackage{multirow}
\usepackage{amssymb}
\usepackage{url}
\usepackage{hhline}
\usepackage[normalem]{ulem}
\useunder{\uline}{\ul}{}

\usepackage{etoolbox}
\makeatletter
\patchcmd{\@makecaption}
  {\scshape}
  {}
  {}
  {}
\makeatletter
\patchcmd{\@makecaption}
  {\\}
  {:\ }
  {}
  {}
\makeatother

\title{\LARGE \bf
   PanopticFusion: Online Volumetric Semantic Mapping \\at the Level of Stuff and Things 
}

\author{
   Gaku Narita, Takashi Seno, Tomoya Ishikawa, Yohsuke Kaji$^{1}$
   \thanks{
      $^{1}$The authors are with R\&D Center, Sony Corporation. 
      {\tt\small \{gaku.narita, takashi.seno, tomoya.ishikawa, yohsuke.kaji\}@sony.com}
   }
}

\begin{document}

\maketitle
\thispagestyle{empty}
\pagestyle{empty}

\begin{abstract}
We propose {\it PanopticFusion}, a novel online volumetric semantic mapping system at the level of {\it stuff} and {\it things}.
In contrast to previous semantic mapping systems, PanopticFusion is able to densely predict class labels of a background region ({\it stuff}) and individually segment arbitrary foreground objects ({\it things}). In addition, our system has the capability to reconstruct a large-scale scene and extract a labeled mesh thanks to its use of a spatially hashed volumetric map representation.
Our system first predicts pixel-wise panoptic labels (class labels for {\it stuff} regions and instance IDs for {\it thing} regions) for incoming RGB frames by fusing 2D semantic and instance segmentation outputs. 
The predicted panoptic labels are integrated into the volumetric map together with depth measurements while keeping the consistency of the instance IDs, which could vary frame to frame, by referring to the 3D map at that moment.
In addition, we construct a fully connected conditional random field (CRF) model with respect to panoptic labels for map regularization. For online CRF inference, we propose a novel unary potential approximation and a map division strategy.

We evaluated the performance of our system on the ScanNet (v2) dataset. PanopticFusion outperformed or compared with state-of-the-art offline 3D DNN methods in both semantic and instance segmentation benchmarks. Also, we demonstrate a promising augmented reality application using a 3D panoptic map generated by the proposed system.
\end{abstract}

\section{INTRODUCTION}
Geometric and semantic scene understanding in 3D environments has an important role in autonomous robotics and context-aware augmented reality (AR) applications.
Geometric scene understanding such as visual simultaneous localization and mapping (SLAM) and 3D reconstruction has been widely discussed since the early days of both the robotics and computer vision communities. 
In recent years, semantic mapping, which not only reconstructs the 3D structure of a scene but also recognizes what exists in the environment, has attracted much attention because of the great progress of deep neural networks.

Semantic mapping systems could take a variety of approaches in terms of geometry and semantics. 
When we think about robotic and AR applications that deeply interact with the real world, what kind of properties are required for the ideal semantic mapping system? 
In terms of geometry, it needs to be able to reconstruct a large-scale scene, not sparsely but densely.
Additionally, the 3D reconstruction desirably needs to be represented as a volumetric map, not just point clouds or surfels, because it is difficult to directly utilize point clouds and surfels for robot--object collision detection or robot navigation.
In terms of semantics, which we mainly focus on in this paper, we believe that it is important for the mapping system to have a {\it holistic} scene understanding capability, that is to say, dense semantic labeling as well as individual object discrimination.
This is because densely labeled semantics is a crucial cue for intelligent robot navigation, and also, discriminating individual objects is essential for robot--object interaction.

\begin{figure}[t]
   \centering
      \includegraphics[height=6.5cm]{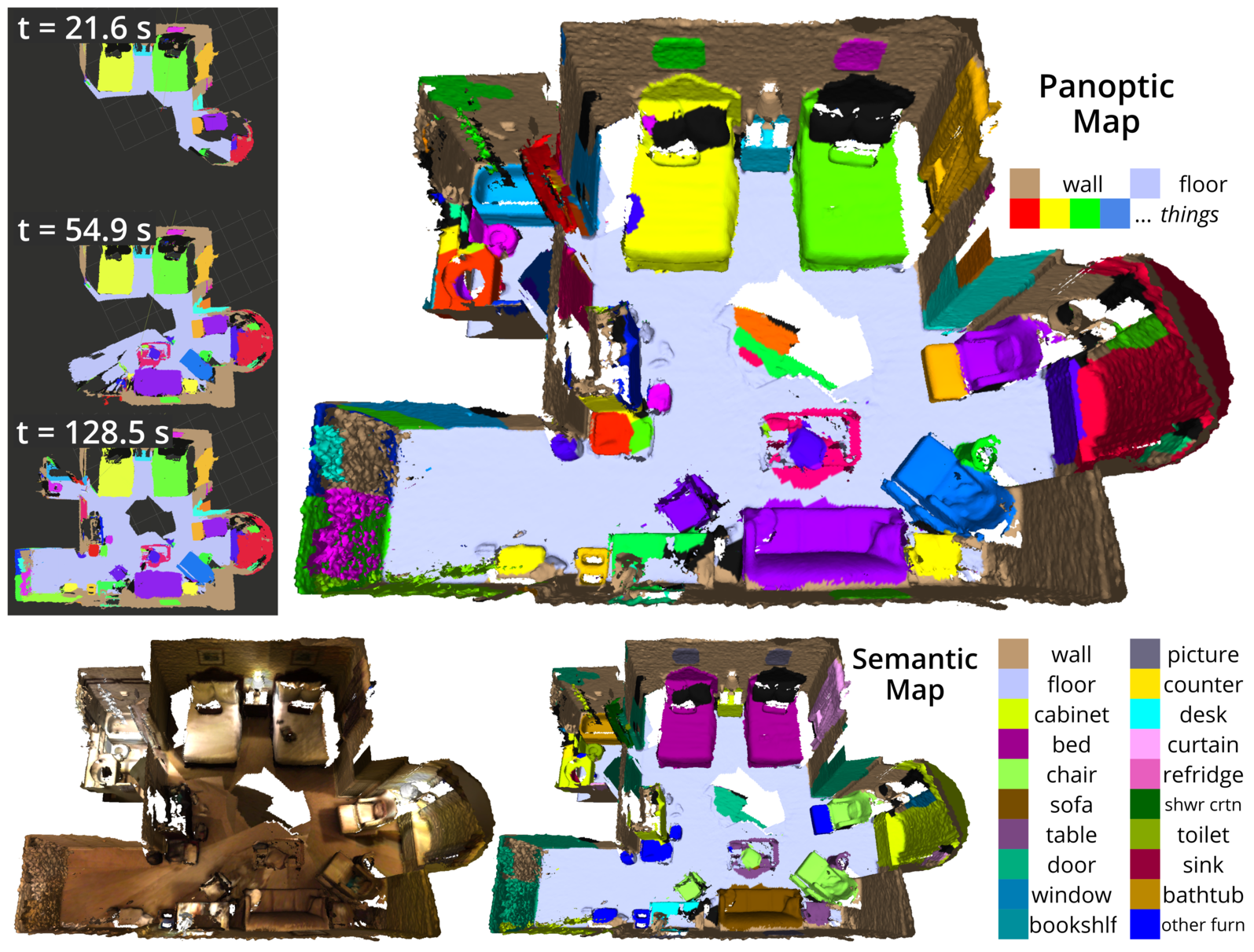}
      \caption{PanopticFusion realizes an online volumetric semantic mapping at the level of {\it stuff} and {\it things}. The system performs large-scale 3D reconstruction, as well as dense semantic labeling on {\it stuff} regions and segmentation of individual {\it things} in an online manner, as shown in the top figure. It is also able to restore the class labels of {\it things} and yield a colored mesh, as shown in the bottom figures. The results obtained with  scene0645\_01 of ScanNet v2 are shown.}
      \label{fig_panopticfusion}
\end{figure}

Turning our eyes to the field of 2D image recognition, an image understanding task called {\it panoptic} segmentation has been proposed recently \cite{kirillov2018panoptic}.
In the panoptic segmentation task, semantic classes are defined as a set of {\it stuff} classes (amorphous regions, such as floors, walls, the sky and roads) and {\it thing} classes (countable objects, such as chairs, tables, people and vehicles) and one needs to predict class labels on {\it stuff} regions and both class labels and instance IDs on {\it thing} regions, where the predictions should be performed for each pixel. 
Extending this point of view to 3D mapping, in this paper we propose the {\it PanopticFusion} system.
To the best of our knowledge, it is the first semantic mapping system that realizes scene understanding at the level of {\it stuff} and {\it things}.
Our system incrementally performs large-scale 3D surface reconstruction online, as well as dense class label prediction on the background region and segmentation and recognition of individual foreground objects, as shown in Fig. \ref{fig_panopticfusion}.

Our approach first passes the incoming RGB frame to 2D semantic and instance segmentation networks and obtains a panoptic label image in which class labels are assigned to {\it stuff} pixels and instance IDs to {\it thing} pixels.
The predicted panoptic labels and depth measurements are integrated into the volumetric map.
Before integration, we keep the consistency of instance IDs, which possibly change from frame to frame, by referring to the volumetric map at that moment.
In addition, we regularize the map using a fully connected CRF model with respect to panoptic labels.
For CRF inference, we propose a unary potential approximation using limited information stored in the map. 
We also present a map division strategy that achieves a significant reduction in computational time without a drop in accuracy.

We evaluated the performance of our system on the ScanNet v2 dataset \cite{dai2017scannet}, a richly annotated large-scale dataset for indoor scene understanding.
The results revealed that PanopticFusion is superior or comparable to the state-of-the-art offline 3D DNN methods in the both 3D semantic and instance segmentation tasks.
Note that our system is not limited to indoor scenes.
Finally, we demonstrated a promising AR application using the 3D panoptic map generated by our system.

The main contributions of this paper are the following:
\begin{itemize}
   \item The first reported semantic mapping system that realizes scene understanding at the level of {\it stuff} and {\it things}.
   \item Large-scale 3D reconstruction and labeled mesh extraction thanks to the use of a spatially hashed volumetric map representation.
   \item Map regularization using a fully connected CRF with a novel unary potential approximation and map division strategy.
   \item Superior or comparable results in both 3D semantic and instance segmentation tasks, in comparison with the state-of-the-art offline 3D DNN methods.
\end{itemize}

\section{RELATED WORK}
Previously proposed representative semantic mapping systems related to our PanopticFusion system are shown in Table \ref{table_related_work}.
These systems can be divided into two categories from the perspective of semantics: the dense labeling approach and the object-oriented approach.

The dense labeling approach builds a single 3D map and assigns a class label or a probability distribution of class labels to each surfel or voxel to realize a dense 3D semantic segmentation.
Hermans {\it et al.} \cite{hermans2014dense} utilize random decision forests for 2D semantic segmentation and transfer the inferred probability distributions to point clouds with a Bayesian update scheme.
Extending the approach of Hermans {\it et al.} \cite{hermans2014dense}, SemanticFusion \cite{mccormac2017semanticfusion} improves the recognition performance by using CNNs for 2D semantic segmentation and makes use of ElasticFusion \cite{Whelan2015ElasticFusionDS} for a SLAM system to generate a globally consistent map.
Xiang {\it et al.} \cite{xiang2017rnn} presented KinectFusion\cite{newcombe2011kinectfusion}-based volumetric mapping with novel data associated RNNs for improving the segmentation accuracy.
While these methods realize dense scene understanding, they suffer from the drawback that they are not able to distinguish individual objects in the scene. 

Methods adopted in the early days of the object-oriented approach leverage 3D model databases.
SLAM++ \cite{salas2013slam++} performs point pair feature-based object detection and feeds the detected objects into a pose graph.
Tateno {\it et al.} \cite{tateno20162} proposed a 3D object detection and pose estimation system that combines unsupervised geometric segmentation and global 3D descriptor matching.
These methods, however, require the shapes of objects in the scene to be exactly the same as the 3D models in the database. 
Recently, several studies on the object-oriented approach using a CNN-based 2D object detector have been reported.
S{\"u}nderhauf {\it et al.} \cite{sunderhauf2017meaningful} and Nakajima {\it et al.} \cite{nakajima2019efficient} combine a 2D object detector and unsupervised geometric segmentation in order to detect objects in point clouds or a surfel map.
MaskFusion \cite{runz2018maskfusion}, Fusion++ \cite{mccormac2018fusion++} and MID-Fusion \cite{xu2018mid} introduced an object-oriented map representation that individually 
builds 3D maps for each object based on 2D object detection.
The object-oriented map representation enables tracking of individual objects \cite{runz2018maskfusion,xu2018mid} and an object-level pose graph optimization \cite{mccormac2018fusion++}.
However, the quantitative recognition performance of these methods is not clear because they mainly evaluate the camera trajectory accuracy.
Furthermore, they focus on foreground objects, resulting in a lack of semantics and/or geometry of background regions.

\begin{table}[t]
   \caption{Semantic mapping systems related to PanopticFusion.}
   \label{table_related_work}
   \centering
   {\scriptsize
      \begin{tabular}{|c|c|c|c|c|c|c|c|c|ccccccccc}
         \cline{1-9}
         \multirow{2}{*}{\textbf{Method}} & \textbf{Speed}            & \multicolumn{3}{c|}{\textbf{Geometry}}                                                                        & \multicolumn{4}{c|}{\textbf{Semantics}}                                                                                                      &  &  &  &  &  &  &  &  &  \\ \cline{2-9}
                                          & \rotatebox{90}{Online }                    & \rotatebox{90}{TSDF Volume} & \rotatebox{90}{Surfels}                   & \rotatebox{90}{Large-scale}               & \rotatebox{90}{Model-free}                & \rotatebox{90}{Dense Labeling } & \rotatebox{90}{Object-level}                 & \rotatebox{90}{Map Reg.}                  &  &  &  &  &  &  &  &  &  \\ \cline{1-9}
         SLAM++ \cite{salas2013slam++}                          & \checkmark &                                                       &                           &                           &                           &                                                          & \checkmark &                           &  &  &  &  &  &  &  &  &  \\ \cline{1-9}
         2.5D is not enough \cite{tateno20162}              & \checkmark & \checkmark                             &                           &                           &                           &                                                          & \checkmark &                           &  &  &  &  &  &  &  &  &  \\ \cline{1-9}
         SemanticFusion \cite{mccormac2017semanticfusion}                  & \checkmark &                                                       & \checkmark & \checkmark & \checkmark & \checkmark                                &                           & \checkmark &  &  &  &  &  &  &  &  &  \\ \cline{1-9}
         DA-RNN \cite{xiang2017rnn}                          & \checkmark & \checkmark                             &                           &                           & \checkmark & \checkmark                                &                           &                           &  &  &  &  &  &  &  &  &  \\ \cline{1-9}
         MaskFusion \cite{runz2018maskfusion}                      & \checkmark &                                                       & \checkmark &                           & \checkmark &                                                          & \checkmark &                           &  &  &  &  &  &  &  &  &  \\ \cline{1-9}
         Fusion++ \cite{mccormac2018fusion++}                        & \checkmark & \checkmark                             &                           &                           & \checkmark &                                                          & \checkmark                          &                           &  &  &  &  &  &  &  &  &  \\ \cline{1-9}
         \textbf{PanopticFusion (Ours)}   & \checkmark & \checkmark                             &                           & \checkmark & \checkmark & \checkmark                                & \checkmark & \checkmark &  &  &  &  &  &  &  &  &  \\ \cline{1-9}
         \end{tabular}
   }
\end{table}

\begin{figure*}[t]
   \centering
      \includegraphics[height=2.38cm]{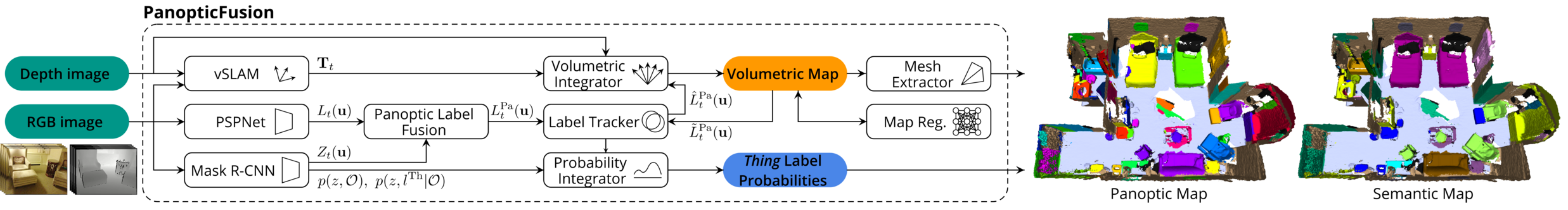}
      \caption{System overview of PanopticFusion.}
      \label{fig_system_overview}
\end{figure*}

In contrast to these related studies, PanopticFusion realizes holistic scene reconstruction and dense semantic labeling with the ability to discriminate individual objects.
Our system builds a single volumetric map, similar to dense labeling approaches, yet each voxel stores neither class labels nor class probability distributions but DNN-predicted panoptic labels in order to seamlessly manage both {\it stuff} and {\it things} semantics.
The class labels of foreground objects can be restored by a probability integration process.
In addition, our 3D reconstruction leverages the truncated signed distance field (TSDF) volumetric map with the voxel hashing data structure \cite{niessner2013real}, which allows us to reconstruct a large-scale scene as well as extract labeled meshes by using marching cubes \cite{lorensen1987marching}, in contrast to the 3D maps of previous methods, which are based on point clouds \cite{hermans2014dense,sunderhauf2017meaningful}, surfels \cite{mccormac2017semanticfusion,nakajima2019efficient,runz2018maskfusion} and a fixed-sized voxel grid \cite{xiang2017rnn,mccormac2018fusion++}.
It should be noted that, with 3D DNN methods that directly apply deep networks to 3D data such as point clouds or voxel grids, high recognition performance has been reported \cite{qi2017pointnet++,dai20183dmv,yi2018gspn,hou20183d}. Nevertheless, with those methods, it is basically necessary to reconstruct the whole scene in advance, requiring offline processing, which could limit their application to robotics and AR. On the contrary, PanopticFusion is an online and incremental framework.

\section{METHOD}
Fig. \ref{fig_system_overview} shows the system overview of PanopticFusion.
Our system first feeds an incoming RGB frame into 2D semantic and instance segmentation networks and obtains pixel-wise panoptic labels by fusing the two outputs (Section \ref{sec_pano_seg}).
The panoptic labels are carefully tracked by referring to the volumetric map at that moment (Section \ref{sec_label_tracking}) and are integrated into the map with depth measurements (Section \ref{sec_volumetric_integration}).
Probability distributions of class labels for foreground objects are also incrementally integrated (Section \ref{sec_label_prob_integration}).
In addition, online map regularization with a fully-connected CRF model is performed for a further improvement of the recognition accuracy.
Note that camera poses with respect to the volumetric map are given by an external vSLAM, and labeled meshes are extracted by using marching cubes \cite{lorensen1987marching}.

\subsection{Notations}
We denote all class labels by $\mathcal{L}$, and they are divided into {\it stuff} labels $\mathcal{L}^\mathrm{St}$ and {\it thing} labels $\mathcal{L}^\mathrm{Th}$:
such that $\mathcal{L} = \mathcal{L}^\mathrm{St} \cup \mathcal{L}^\mathrm{Th}$ and $\mathcal{L}^\mathrm{St} \cap \mathcal{L}^\mathrm{Th} = \emptyset$.
A set of instance IDs for discriminating individual {\it things} is denoted by $\mathcal{Z}$.
Here we define a set of panoptic labels $\mathcal{L}^\mathrm{Pa} = \mathcal{L}^\mathrm{St} \cup \mathcal{Z} \cup l_\mathrm{unk}$ in order to seamlessly manage {\it stuff} and {\it things} level semantics in the 3D map.
$l_\mathrm{unk}$ denotes the {\it unknown} label.

\subsection{Volumetric Map}
We use the TSDF-based volumetric map representation with a voxel hashing approach \cite{niessner2013real}, which manages spatially hashed small regular voxel grids called voxel blocks. 
This approach is memory efficient compared with a single voxel grid approach like the original KinectFusion \cite{newcombe2011kinectfusion} and enables us to reconstruct large-scale scenes.
Our implementation is based on voxblox \cite{oleynikova2017voxblox}, which is a CPU-based TSDF mapping system, but we extend it to integrate the semantics. 

Our volumetric map stores the truncated signed distance $\mathtt{D}_t(\mathbf{v}) \in \mathbb{R}$, the RGB color $\mathtt{C}_t(\mathbf{v}) \in \mathbb{R}^3$ and the associated weight $\mathtt{W}_t^\mathrm{D}(\mathbf{v}) \in \mathbb{R}_{\geq 0}$ at each voxel location $\mathbf{v} \in \mathbb{R}^3$, as with \cite{newcombe2011kinectfusion}.
Our system additionally stores the panoptic label $\mathtt{L}_t^\mathrm{Pa}(\mathbf{v}) \in \mathcal{L}^\mathrm{Pa}$ and its weight $\mathtt{W}_t^\mathrm{L}(\mathbf{v}) \in \mathbb{R}_{\geq 0}$.
Here $t$ denotes the time index.

\subsection{2D Panoptic Label Prediction\label{sec_pano_seg}}
For the incoming RGB frame, we predict pixel-wise panoptic labels by fusing both 2D semantic and instance segmentation outputs.
We utilize the state-of-the-art CNN architectures of PSPNet \cite{zhao2017pyramid} and Mask R-CNN \cite{he2017mask} for 2D semantic and instance segmentation, respectively.
PSPNet infers pixel-wise class labels $L_t(\mathbf{u}) \in \mathcal{L}$, where $\mathbf{u} \in \mathbb{R}^2$ denotes the image coordinates.
Mask R-CNN outputs instance IDs for each pixel $Z_t(\mathbf{u}) \in \mathcal{Z} \cup l_\mathrm{unk}$, where the regions without any foreground objects are filled with $l_\mathrm{unk}$.
The foreground object probability $p_t(z, \mathcal{O})$ and conditional probability distribution of {\it thing} labels $p_t(z, l^\mathrm{Th} | \mathcal{O})$ with respect to instance $z$ are utilized in the probability integration step described in Section \ref{sec_label_prob_integration}.
We obtain pixel-wise panoptic labels $L_t^\mathrm{Pa}(\mathbf{u})$ from $L_t(\mathbf{u})$ and $Z_t(\mathbf{u})$ preceding the instance IDs:
\begin{eqnarray}
   L_t^\mathrm{Pa}(\mathbf{u}) = \begin{cases}
      Z_t(\mathbf{u}) & Z_t(\mathbf{u}) \neq l_\mathrm{unk} \\
      L_t(\mathbf{u}) & Z_t(\mathbf{u}) = l_\mathrm{unk} \land L_t(\mathbf{u}) \in \mathcal{L}^\mathrm{St} \\
      l_\mathrm{unk} & \mathrm{otherwise.}
   \end{cases}
\end{eqnarray}

\subsection{Panoptic Label Tracking\label{sec_label_tracking}}
Direct integration of raw panoptic labels $L_t^\mathrm{Pa}(\mathbf{u})$ into the volumetric map induces label inconsistency because Mask R-CNN does not necessarily output a consistent instance ID for the same object through multiple frames.
To avoid this problem, we need to estimate consistency-resolved panoptic labels $\hat{L}_t^\mathrm{Pa}(\mathbf{u})$ before the integration.
The simplest way is to track the foreground objects in the 2D image sequence using a visual object tracker.
This approach unfortunately is not able to re-identify an object in the case of a loopy camera trajectory.
Therefore, we take a map reference approach similar to \cite{runz2018maskfusion,mccormac2018fusion++}.

We first prepare the reference panoptic labels $\tilde{L}_{t-1}^\mathrm{Pa}(\mathbf{u})$ by accessing the  map. 
Here, $\mathbf{T}_{t}$ denotes the live camera pose, $\mathbf{K}$ the camera intrinsic parameters, and $D_t(\mathbf{u})$ the live depth map:
\begin{eqnarray}
   \tilde{L}_{t-1}^\mathrm{Pa}(\mathbf{u}) = \mathtt{L}_{t-1}^\mathrm{Pa}(\mathbf{T}_{t} \mathbf{K}^{-1} D_t(\mathbf{u}) [\mathbf{u}, 1]^\mathrm{T}).
\end{eqnarray}
To track labels, we compute the intersection over union (IoU) $U(\tilde{z}, z)$ of instance ID $z$ of raw panoptic labels $L_t^\mathrm{Pa}(\mathbf{u})$ and instance ID $\tilde{z}$ of reference panoptic labels $\tilde{L}_{t-1}^\mathrm{Pa}(\mathbf{u})$:
\begin{eqnarray}
   U(\tilde{z}, z) = \mathrm{IoU}\bigr(\{\mathbf{u} | \tilde{L}_{t-1}^\mathrm{Pa}(\mathbf{u}) = \tilde{z}\}, \{\mathbf{u} | L_t^\mathrm{Pa}(\mathbf{u}) = z\}\bigl)
\end{eqnarray}
Here, IoU is defined as $\mathrm{IoU}(A,B) = |A \cap B| / |A \cup B|$.

When the maximum value of IoU exceeds a threshold $\theta_U$, $\tilde{z}$ giving the maximum value is associated with $z$. Otherwise a new instance ID is assigned to $z$:
\begin{eqnarray}
   \hat{z} = \begin{cases}
      \mathrm{arg}\max_{\tilde{z}} U(\tilde{z}, z) & \max_{\tilde{z}} U(\tilde{z}, z) > \theta_U \\
      z_\mathrm{new} & \mathrm{otherwise.}
   \end{cases} 
\end{eqnarray}
The association is processed in descending order in the mask area $|\{\mathbf{u} | L_t^\mathrm{Pa}(\mathbf{u}) = z\}|$. Once a reference instance ID $\tilde{z}$ is associated with $z$, that instance ID is not associated with any other $z$.
The utilization of IoU instead of an overlap ratio, as used in  \cite{runz2018maskfusion,mccormac2018fusion++}, and the exclusive label association is for avoiding under-segmentation of foreground objects in the map.
From the associated instance IDs and raw {\it stuff} labels, we obtain the consistency-resolved panoptic labels $\hat{L}_t^\mathrm{Pa}(\mathbf{u})$ as follows, which are used in the integration step:
\begin{eqnarray}
   \hat{L}_t^\mathrm{Pa}(\mathbf{u}) = \begin{cases}
      L_t^\mathrm{Pa}(\mathbf{u}) & L_t^\mathrm{Pa}(\mathbf{u}) \in \mathcal{L}^\mathrm{St} \\
      \hat{z} & L_t^\mathrm{Pa}(\mathbf{u}) \in \mathcal{Z} \\
      l_\mathrm{unk} & \mathrm{otherwise.}   
   \end{cases} 
\end{eqnarray}

\subsection{Volumetric Integration\label{sec_volumetric_integration}}
For integration, we take the raycasting approach, as with \cite{oleynikova2017voxblox}.
For each pixel $\mathbf{u}$, we cast a ray from the sensor origin $\mathbf{s}$ to the back-projected 3D point $\mathbf{p}_\mathbf{u} = \mathbf{T}_{t} \mathbf{K}^{-1} D_t(\mathbf{u}) [\mathbf{u}, 1]^\mathrm{T}$ and update the voxels along the ray within a truncated distance.
Regarding TSDF values, we update them by weighted averaging, similar to \cite{newcombe2011kinectfusion}:
\begin{align}
   \mathtt{D}_t(\mathbf{v}) &= \frac{\mathtt{W}_{t-1}^\mathrm{D}(\mathbf{v}) \mathtt{D}_{t-1}(\mathbf{v}) + w_t(\mathbf{v}, \mathbf{p}_\mathbf{u}) d_t(\mathbf{v}, \mathbf{p}_\mathbf{u}, \mathbf{s})}{\mathtt{W}_{t-1}^\mathrm{D}(\mathbf{v}) + w_t(\mathbf{v}, \mathbf{p}_\mathbf{u})}, \\
   \mathtt{W}_t^\mathrm{D}(\mathbf{v}) &= \mathtt{W}_{t-1}^\mathrm{D}(\mathbf{v}) + w_t(\mathbf{v}, \mathbf{p}_\mathbf{u}). \label{eq_tsdf_integ}
\end{align}
Here, $d_t$ denotes the distance between the voxel and the surface boundary, and $w_t$ a quadric weight \cite{oleynikova2017voxblox} that takes the reliability of depth measurements into account.
Similar updating is applied to the voxel color $\mathtt{C}_t(\mathbf{v})$.

In contrast to TSDF and colors of continuous values,  weighted averaging cannot be applied to panoptic labels of discrete values.
The most reliable and simplest way to manage panoptic labels is to record all integrated labels.
This, unfortunately, results in a significant increase in memory usage and frequent memory allocation.
Instead we store a single label at each voxel and update its weight by the increment/decrement strategy.
If the pixel-wise panoptic label $\hat{L}_t^\mathrm{Pa}(\mathbf{u})$ estimated in the previous section is the same as the current voxel panoptic label $\mathtt{L}_{t-1}^\mathrm{Pa}(\mathbf{v})$, we increment the weight $\mathtt{W}_t^\mathrm{L}(\mathbf{v})$ with the quadric weight:
\begin{eqnarray}
   \mathtt{L}_t^\mathrm{Pa}(\mathbf{v}) = \mathtt{L}_{t-1}^\mathrm{Pa}(\mathbf{v}), \ 
   \mathtt{W}_t^\mathrm{L}(\mathbf{v}) = \mathtt{W}_{t-1}^\mathrm{L}(\mathbf{v}) + w_t(\mathbf{v}, \mathbf{p}_\mathbf{u}). \label{eq_label_integ_case1}
\end{eqnarray}
In contrast, if those panoptic labels do not coincide, we decrement the weight:
\begin{eqnarray}
   \mathtt{L}_t^\mathrm{Pa}(\mathbf{v}) = \mathtt{L}_{t-1}^\mathrm{Pa}(\mathbf{v}), \ 
   \mathtt{W}_t^\mathrm{L}(\mathbf{v}) = \mathtt{W}_{t-1}^\mathrm{L}(\mathbf{v}) - w_t(\mathbf{v}, \mathbf{p}_\mathbf{u}). \label{eq_label_integ_case2}
\end{eqnarray}
Note that in the case where $w_t > \mathtt{W}_{t-1}^\mathrm{L}$, that is, when the weight considerably falls, we replace the voxel label with the newly estimated label:
\begin{eqnarray}
   \mathtt{L}_t^\mathrm{Pa}(\mathbf{v}) = \hat{L}_t^\mathrm{Pa}(\mathbf{u}), \ 
   \mathtt{W}_t^\mathrm{L}(\mathbf{v}) = w_t(\mathbf{v}, \mathbf{p}_\mathbf{u}) - \mathtt{W}_{t-1}^\mathrm{L}(\mathbf{v}). \label{eq_label_integ_case3}
\end{eqnarray}

\subsection{Thing Label Probability Integration\label{sec_label_prob_integration}}
The {\it thing} label predicted by Mask R-CNN is frequently uncertain even while the segmentation mask is accurate, especially in the case where a small part of the object is visible.
Hence we probabilistically integrate {\it thing} labels instead of assigning a single label to each foreground object:
\begin{eqnarray}
   p_{1 \cdots t}(z, l^\mathrm{Th}) = \frac{\sum_t p_t(z, \mathcal{O}) p_t(z, l^\mathrm{Th}|\mathcal{O})}{\sum_t p_t(z, \mathcal{O})}.
\end{eqnarray}
Weighting the probability distributions with the detection confidence $p_t(z, \mathcal{O})$ allows the final distribution to preferentially reflect reliable detections.

\subsection{Online Map Regularization\label{sec_map_regularization}}
While the integration scheme described above yields a reliable 3D panoptic map, it is possible to further improve the recognition accuracy by using a map regularization with a fully connected CRF model.
A fully connected CRF with Gaussian edge potentials has been widely used in 2D image segmentation since an efficient inference method was proposed \cite{krahenbuhl2011efficient}.
Recently, several studies that apply it to a 3D map, such as surfels or occupancy grids, have been reported \cite{hermans2014dense,mccormac2017semanticfusion,yang2017}.
In those approaches, CRF models are constructed with respect to class labels whose number is fixed,
whereas we consider the CRF with respect to panoptic labels whose number depends on the scene and is theoretically not limited.
Here we are faced with two problems:
how to properly compute unary potentials for panoptic labels, and how to infer a CRF whose number of labels is potentially large within a practical time.

\subsubsection{Problem Setting}
We construct a fully connected graph whose nodes are individual voxels.
We assign a label variable $x_v \in \mathcal{L}^\mathrm{Pa}$ to each node and infer the optimal labels $\mathbf{x} = \{x_v\}$ that minimize the Gibbs energy $E$ by the mean-field approximation and a message passing scheme:
\begin{eqnarray}
   E(\mathbf{x}) = \sum_v \psi_u(x_v) + \sum_{v<v^\prime} \psi_p(x_v, x_{v^\prime}).
\end{eqnarray}
While it is non-trivial which unary potentials should be used for a panoptic label CRF, we use a negative logarithm of a probability distribution following a standard class label CRF:
\begin{eqnarray}
   \psi_u(x_v) = -\log p(x_v). \label{eq_unary_potential}
\end{eqnarray}
We utilize a linear combination of Gaussian kernels for pairwise potentials because the efficient inference method  \cite{krahenbuhl2011efficient} can be applied:
\begin{eqnarray}
   \psi_p(x_v, x_{v^\prime}) = \mu(x_v, x_{v^\prime}) \sum_m w^{(m)} k^{(m)}(\mathbf{f}_v, \mathbf{f}_{v^\prime}).
\end{eqnarray}
Here, $\mu(x_s, x_s^\prime) = 1_{[x_s \neq x_s^\prime]}$ is a simple Potts model.
As in \cite{krahenbuhl2011efficient}, we chose the following two kernels which regularize the map with respect to voxel colors and locations, respectively:
\begin{align}
   k^{(1)}(\mathbf{f}_v, \mathbf{f}_{v^\prime}) &= \exp\biggr(-\frac{|\mathbf{v} - \mathbf{v}^{\prime}|^2}{2 \theta_\alpha^2} - \frac{|\mathtt{C}(\mathbf{v}) - \mathtt{C}(\mathbf{v}^{\prime})|^2}{2 \theta_\beta^2}\biggl), \\
   k^{(2)}(\mathbf{f}_v, \mathbf{f}_{v^\prime}) &= \exp\biggr(-\frac{|\mathbf{v} - \mathbf{v}^{\prime}|^2}{2 \theta_\alpha^2}\biggl).
\end{align}

\subsubsection{Unary Potential Approximation}
Previous approaches \cite{hermans2014dense,mccormac2017semanticfusion,yang2017} assigned a probability distribution to each surfel or voxel, which can be used directly to compute unary potentials; 
in contrast, from the viewpoint of memory efficiency, we store only a single label in each voxel.
Therefore, we approximate the unary potentials using only a single label, and weights stored in a voxel, based on a certain assumption described as follows.

Here let us focus on the integration scheme of panoptic labels shown in Eq. \eqref{eq_label_integ_case1}-\eqref{eq_label_integ_case3}.
We denote the set of times when the predicted panoptic label is the same as, and not the same as, the current voxel label by $\mathcal{T}_{+} = \{\tau \ | \ \hat{L}_\tau^\mathrm{Pa}(\mathbf{u}) = \mathtt{L}_t^\mathrm{Pa}(\mathbf{v})\}$ and $\mathcal{T}_{-} = \{\tau \ | \ \hat{L}_\tau^\mathrm{Pa}(\mathbf{u}) \neq \mathtt{L}_t^\mathrm{Pa}(\mathbf{v})\}$, respectively.
If $\mathtt{L}_\tau^\mathrm{Pa}(\mathbf{v}) = \mathtt{L}_t^\mathrm{Pa}(\mathbf{v})$ for all $\tau = 1, \cdots, {t-1}$, that is to say, the voxel label has not changed, Eq. \eqref{eq_label_weight_approx} holds strictly.
If $p(x_v = \mathtt{L}_t^\mathrm{Pa}(\mathbf{v})) > 0.5$ and the number of integrations is sufficiently large, Eq. \eqref{eq_label_weight_approx} holds asymptotically:
\begin{eqnarray}
   \sum_{t \in \mathcal{T}_{+}} w_t(\mathbf{v}, \mathbf{p}_\mathbf{u}) - \sum_{t \in \mathcal{T}_{-}} w_t(\mathbf{v}, \mathbf{p}_\mathbf{u}) \simeq \mathtt{W}_t^\mathrm{L}(\mathbf{v}). \label{eq_label_weight_approx}
\end{eqnarray}
In addition, from the TSDF update scheme in Eq. \eqref{eq_tsdf_integ} we have,
\begin{eqnarray}
   \sum_{t \in \mathcal{T}_{+}} w_t(\mathbf{v}, \mathbf{p}_\mathbf{u}) + \sum_{t \in \mathcal{T}_{-}} w_t(\mathbf{v}, \mathbf{p}_\mathbf{u}) = \mathtt{W}_t^\mathrm{D}(\mathbf{v}).
\end{eqnarray}
Consequently, the probability that the current panoptic label in the voxel is actually correct can be calculated as,
\begin{align}
   p(x_v = \mathtt{L}_t^\mathrm{Pa}(\mathbf{v})) &= \frac{\sum_{t \in \mathcal{T}_{+}} w_t(\mathbf{v}, \mathbf{p}_\mathbf{u})}{\sum_{t \in \mathcal{T}_{+}} w_t(\mathbf{v}, \mathbf{p}_\mathbf{u}) + \sum_{t \in \mathcal{T}_{-}} w_t(\mathbf{v}, \mathbf{p}_\mathbf{u})} \nonumber \\
   &\simeq \frac{1}{2} \biggl(1 + \frac{\mathtt{W}_t^\mathrm{L}(\mathbf{v})}{\mathtt{W}_t^\mathrm{D}(\mathbf{v})}\biggr). \label{eq_approx_positive_prob}
\end{align}
It is unfortunately not possible to calculate the exact probability that the voxel takes a label other than the current label because the map does not record all the information about previously integrated labels.
Therefore, we approximate the probability as follows, where $M$ denotes the number of panoptic labels in the map:
\begin{eqnarray}
   p(x_v) = \frac{1}{M-1} \bigl(1 - p(x_v = \mathtt{L}_t^\mathrm{Pa}(\mathbf{v}))\bigr) \ \ (x_v \neq \mathtt{L}_t^\mathrm{Pa}(\mathbf{v})). \label{eq_approx_negative_prob}
\end{eqnarray}

Finally, we obtain the unary potential from Eq. \eqref{eq_unary_potential}, \eqref{eq_approx_positive_prob} and \eqref{eq_approx_negative_prob}.
In spite of the approximated approach, it realizes quantitative and qualitative improvements in recognition accuracy, as shown in Section \ref{sec_eval_crf}.

\subsubsection{Map Division for Online Inference}
The computational complexity of the inference algorithm proposed by Kr{\"a}henb{\"u}hl {\it et al.} \cite{krahenbuhl2011efficient} is $\mathcal{O}(NM)$, where $N$ and $M$ are the numbers of voxels and panoptic labels, respectively.
In our problem setting, however, $M$ is theoretically limitless and could in practice be large, e.g. several hundreds, which would make online inference impracticable.
To solve this problem, we present a map division strategy.
When we divide the volumetric map into $S$ spatially contiguous submaps, the number of panoptic labels in each submap can be expected to be $\mathcal{O}(M/S)$.
Hence, the total computational complexity could be reduced to $S \times \mathcal{O}(N/S \times M/S) = \mathcal{O}(NM) / S$.
The map is divided by the block-wise region growing approach based on the predefined maximum number of voxel blocks. The division process has little effect on computational time.

\section{EVALUATION}
\subsection{Experimental Setup}
\begin{figure*}[t]
   \centering
      \includegraphics[height=7cm]{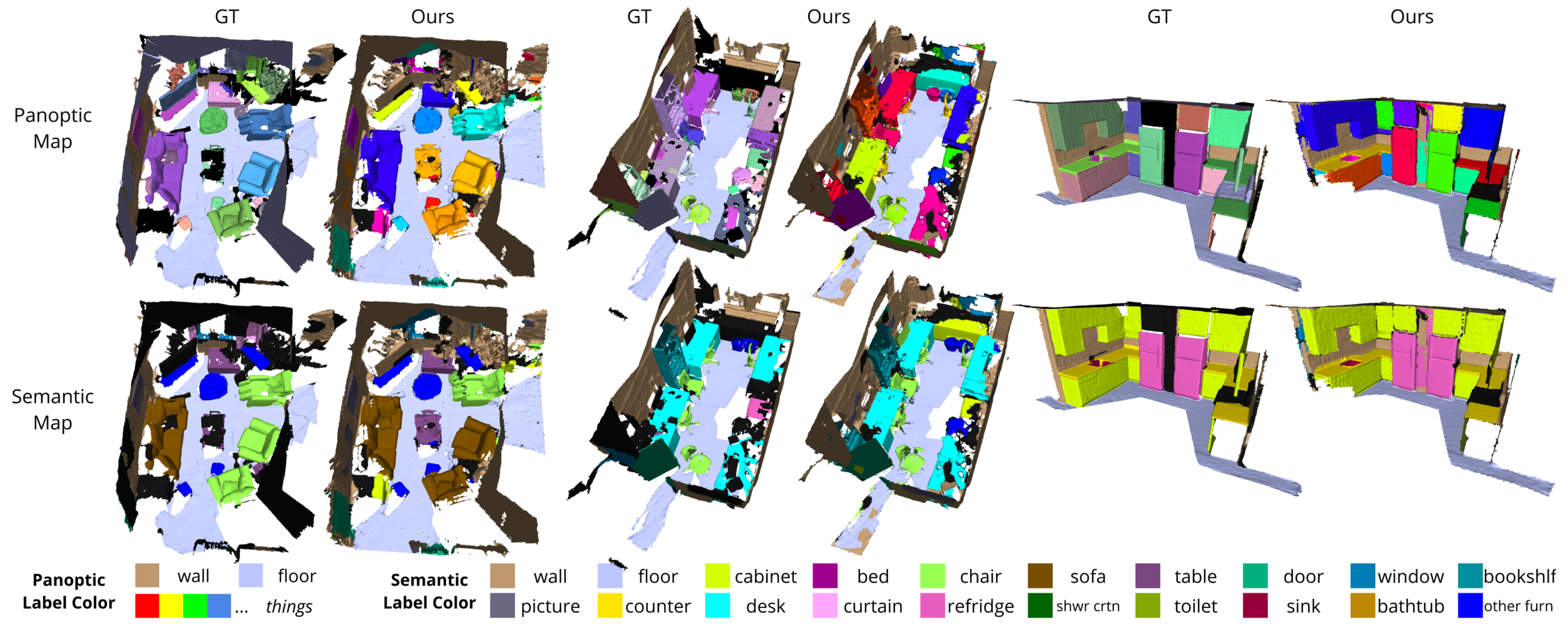}
      \caption{Qualitative results obtained with PanopticFusion system. From left to right, typical scenes in ScanNet v2 of scene0608\_00, scene0643\_00 and scene0488\_01 are displayed. Note that ground truth and our results leverage different reconstruction algorithms, and the colors of {\it things} in our results are not necessarily the same as the ground truth.}
      \label{fig_qualitative_results}
\end{figure*}

For evaluating the performance of our system, we used the ScanNet v2 dataset \cite{dai2017scannet}, 
a large-scale dataset for indoor scene understanding.
It provides RGB-D images captured by hand-held consumer-grade depth sensors, camera trajectories, reconstructed 3D models, and 2D/3D semantic annotations.
In the following experiments, we used RGB-D images of size 640$\times$480 pixels and the provided camera trajectories for fair comparison.
The dataset was composed of 1201 training scenes and 312 {\it open} test scenes.
In addition, 100 {\it hidden} test scenes without publicly available semantic annotations are provided for the ScanNet Benchmark Challenge \cite{scannet_benchmark_challenge}. 
For quantitative evaluations, 20 class annotations are generally used.
In this paper, we define the wall and floor as the {\it stuff} class $\mathcal{L}^\mathrm{St}$ and the other 18 classes, such as chairs and sofas, as the {\it thing} class $\mathcal{L}^\mathrm{Th}$. 
Note that our system is not limited to indoor scenes, and the numbers of {\it stuff} and {\it thing} classes can be arbitrarily defined.

We employed ResNet-50 for the backbone of PSPNet. The network was initialized with the ADE20K \cite{zhou2016semantic} pre-trained weights, and was then fine-tuned using a SGD optimizer for 30 epochs with a learning rate of 0.01 and a batch size of 2.
We leveraged ResNet-101-FPN for the Mask R-CNN's backbone. After initialization with MS COCO \cite{lin2014microsoft} pre-trained weights, the network was fine-tuned by 4-step alternating learning \cite{ren2015faster} using an ADAM optimizer for 25 epochs with a learning rate of 0.001 and a batch size of 1\footnote{We used a publicly available implementation of \cite{pspnet_keras_tensorflow} and \cite{matterport_maskrcnn_2017} for PSPNet and Mask R-CNN, respectively.}.

We used the following parameters for the integration process: voxel size of 0.024 m, a truncation distance of 4$\times$0.024 m, 16$\times$16$\times$16 voxels per voxel block, IoU threshold $\theta_U = 0.25$.
In the map regularization, $w^{(1)} = 10$, $w^{(2)} = 15$, $\theta_\alpha = 0.05 \ \mathrm{m}$, $\theta_\beta = 20$ were used with 5 iterations of CRF inference.
The following experiments were performed on a computer equipped with an Intel Core i7-7800X CPU at 3.50 GHz and two NVIDIA GeForce GTX 1080Ti GPUs.

\subsection{Quantitative and Qualitative Results}
Fig. \ref{fig_qualitative_results} shows examples of 3D panoptic maps generated by our system.
Unfortunately, there are no semantic mapping systems or 3D DNNs that can recognize a 3D scene at the level of {\it stuff} and {\it things}.
Therefore, we evaluated the performance on two sub-tasks, 3D semantic segmentation and instance segmentation, for a quantitative comparison.
In this evaluation, we used the hidden test set of ScanNet v2. 
We show the results in Tables \ref{table_scannet_sema_seg_benchmark} and \ref{table_scannet_inst_seg_benchmark}.
In the tables, the state-of-the-art methods that apply 3D DNNs to points or volumetric grids are listed.
Note that the methods of \cite{huang2018texturenet,dai20183dmv,hou20183d} leverage RGB images with associated camera poses as well.
Our system that uses only 2D-based recognition modules surprisingly achieves comparable or superior performance compared with those methods, thanks to the careful integration of multi-view predictions.
In terms of the class-wise accuracy, the results revealed that our system has advantages especially in the case of small objects such as sinks and pictures, and objects that are confusing to recognize only by their geometry, such as beds, bookshelves, and curtains.
In Table \ref{table_scannet_sema_seg_benchmark}, several semantic segmentation methods outperform our system because of their large receptive fields in 3D space. However, these methods basically need to reconstruct the entire scene in advance, assuming offline process, while our system is an online and incremental framework.
How to apply 3D DNNs to partial observations and how to integrate them into an online mapping system are left for future work.

Additionally, we evaluated 3D panoptic quality on the open test set of ScanNet v2, although there are no quantitatively comparable methods.
We employed the evaluation criteria originally proposed in \cite{kirillov2018panoptic}.
Note that the quality was evaluated with respect to each vertex instead of each pixel, and, as with the ScanNet 3D semantic instance benchmark, we ignored the predicted {\it things} with less than 100 vertices.
We show the panoptic quality (PQ) as well as the segmentation quality (SQ) and recognition quality (RQ) in Table \ref{table_panoptic_quality}.
We hope these results will invigorate research in this field.

\begin{table*}[t]
   \renewcommand{\baselinestretch}{0.8}
   \caption{3D semantic segmentation results on ScanNet (v2) 3D semantic label benchmark (hidden test set) \cite{scannet_benchmark_challenge}. This table shows IoU (\%). Note that the bold and underlined numbers denote first and second ranks, respectively.}
   \label{table_scannet_sema_seg_benchmark}
   \centering
   \scalebox{0.8}[0.8]{
      \begin{tabular}{c||c|p{1.4em}p{1.4em}p{1.4em}p{1.4em}p{1.4em}p{1.4em}p{1.4em}p{1.4em}p{1.4em}p{1.4em}p{1.4em}p{1.4em}p{1.4em}p{1.4em}p{1.4em}p{1.4em}p{1.4em}p{1.4em}p{1.4em}p{1.4em}}
      \hline

                            & avg. & wall & floor & cab  & bed  & chair & sofa & tabl & door & wind & bkshf & pic  & cntr & desk & curt & fridg & showr & toil & sink & bath & ofurn \\ \hline
      ScanNet \cite{dai2017scannet}              & 30.6 & 43.7 & 78.6  & 31.1 & 36.6 & 52.4  & 34.8 & 30.0 & 18.9 & 18.2 & 50.1  & 10.2 & 21.1 & 34.2 & 0.2  & 24.5  & 15.2  & 46.0 & 31.8 & 20.3 & 14.5  \\ 
      PointNet++ \cite{qi2017pointnet++}           & 33.9 & 52.3 & 67.7  & 25.6 & 47.8 & 36.0  & 34.6 & 23.2 & 26.1 & 25.2 & 45.8  & 11.7 & 25.0 & 27.8 & 24.7 & 21.2  & 14.5  & 54.8 & 36.4 & 58.4 & 18.3  \\
      SPLATNet \cite{su2018splatnet}             & 39.3 & {\ul 69.9} & 92.7 & 31.1 & 51.1 & 65.6 & 51.0 & 38.3 & 19.7 & 26.7 & 60.6 & 0.0 & 24.5 & 32.8 & 40.5 & 0.1 & 24.9 & 59.3 & 27.1 & 47.2 & 22.7  \\
      Tangent Conv. \cite{tatarchenko2018tangent}  & 43.8 & 63.3 & 91.8  & 36.9 & 64.6 & 64.5  & 56.2 & 42.7 & 27.9 & 35.2 & 47.4  & 14.7 & 35.3 & 28.2 & 25.8 & 28.3  & 29.4  & 61.9 & 48.7 & 43.7 & 29.8  \\
      3DMV \cite{dai20183dmv}                 & 48.4 & 60.2 & 79.6 & 42.4 & 53.8 & 60.6 & 50.7 & 41.3 & 37.8 & 53.9 & 64.3 & 21.4 & 31.0 & 43.3 & 57.4 & {\ul 53.7} & 20.8 & 69.3 & 47.2 & 48.4 & 30.1  \\
      TextureNet \cite{huang2018texturenet}           & {\ul 56.6} & 68.0 & {\ul 93.5} & {\ul 49.4} & 66.4 & {\ul 71.9} & 63.6 & {\ul 46.4} & {\ul 39.6} & {\ul 56.8} & {\ul 67.1} & 22.5 & {\ul 44.5} & 41.1 & 67.8 & 41.2 & 53.5 & 79.4 & {\ul 56.5} & \textbf{67.2} & {\ul 35.6}  \\
      SparseConvNet \cite{graham20183d}        & \textbf{72.5} & \textbf{86.5} & \textbf{95.5} & \textbf{72.1} & \textbf{82.1} & \textbf{86.9} & \textbf{82.3} & \textbf{62.8} & \textbf{61.4} & \textbf{68.3} & \textbf{84.6} & \textbf{32.5} & \textbf{53.3} & \textbf{60.3} & \textbf{75.4} & \textbf{71.0} & \textbf{87.0} & \textbf{93.4} & \textbf{72.4} & {\ul 64.7} & \textbf{57.2}  \\ \hline
      \textbf{PanopticFusion (Ours)}  & 52.9 & 60.2 & 81.5 & 38.6 & {\ul 68.8} & 63.2 & {\ul 64.9} & 44.2 & 29.3 & 56.1 & 60.4 & {\ul 24.1} & 22.5 & {\ul 43.4} & {\ul 70.5} & 49.9 & {\ul 66.9} & {\ul 79.6} & 50.7 & 49.1 & 34.8 \\ \hline
   \end{tabular}
   }
\end{table*}

\begin{table*}[t]
   \renewcommand{\baselinestretch}{0.8}
   \caption{3D instance segmentation results on ScanNet (v2) 3D semantic instance benchmark (hidden test set) \cite{scannet_benchmark_challenge}. This table shows AP$_{0.5}$, average precision with IoU threshold of 0.5. Note that the bold and underlined numbers denote first and second ranks, respectively.}
   \label{table_scannet_inst_seg_benchmark}
   \centering
   \scalebox{0.8}[0.8]{
      \begin{tabular}{c||c|p{1.4em}p{1.4em}p{1.4em}p{1.4em}p{1.4em}p{1.4em}p{1.4em}p{1.4em}p{1.4em}p{1.4em}p{1.4em}p{1.4em}p{1.4em}p{1.4em}p{1.4em}p{1.4em}p{1.4em}p{1.4em}}
      \hline
                                    & avg. & cab  & bed  & chair & sofa & tabl & door & wind & bkshf & pic  & cntr & desk & curt & fridg & showr & toil & sink & bath & ofurn \\ \hline
      SGPN \cite{wang2018sgpn}      & 14.3 & 6.5 & 39.0 & 27.5 & 35.1 & 16.8 & 8.7 & 13.8 & 16.9 & 1.4 & {\ul 2.9} & 0.0 & 6.9 & 2.7 & 0.0 & 43.8 & 11.2 & 20.8 & 4.3   \\
      GSPN \cite{yi2018gspn}  & 30.6 & {\ul 34.8} & 40.5 & {\ul 58.9} & 39.6 & {\ul 27.5} & 28.3 & 24.5 & 31.1 & 2.8 & \textbf{5.4} & 12.6 & 6.8 & 21.9 & 21.4 & 82.1 & 33.1 & 50.0 & 29.0  \\ 
      3D-SIS \cite{hou20183d}       & 38.2 & 19.0 & 43.2 & 57.7 & \textbf{69.9} & 27.1 & {\ul 32.0} & 23.5 & 24.5 & 7.5 & 1.3 & 3.3 & 26.3 & \textbf{42.2} & \textbf{85.7} & 88.3 & 11.7 & \textbf{100.0} & 24.0  \\ 
      MASC \cite{liu2019masc}                         & {\ul 44.7} & \textbf{38.2} & {\ul 55.5} & \textbf{63.3} & {\ul 63.9} & \textbf{38.6} & \textbf{36.1} & {\ul 27.6} & {\ul 38.1} & {\ul 32.7} & 0.2 & \textbf{26.0} & {\ul 50.9} & {\ul 45.1} & 57.1 & \textbf{98.0} & {\ul 36.7} & 52.8 & {\ul 43.2}  \\ \hline
      \textbf{PanopticFusion (Ours)}         & \textbf{47.8} & 25.9 & \textbf{71.2} & 55.0 & 59.1 & 26.7 & 25.0 & \textbf{35.9} & \textbf{59.5} & \textbf{43.7} & 0.0 & {\ul 17.5} & \textbf{61.3} & 41.1 & \textbf{85.7} & {\ul 94.4} & \textbf{48.5} & {\ul 66.7} & \textbf{43.4}  \\ \hline
      \end{tabular}
   }
\end{table*}

\begin{table*}[t]
   \renewcommand{\baselinestretch}{0.8}
   \caption{3D panoptic segmentation results on ScanNet (v2) open test set.}
   \label{table_panoptic_quality}
   \centering
   \scalebox{0.75}[0.75]{
      \begin{tabular}{c|c||c|cc|p{1.4em}p{1.4em}p{1.4em}p{1.4em}p{1.4em}p{1.4em}p{1.4em}p{1.4em}p{1.4em}p{1.4em}p{1.4em}p{1.4em}p{1.4em}p{1.4em}p{1.4em}p{1.4em}p{1.4em}p{1.4em}p{1.4em}p{1.4em}}
         \hline
         method & metric & all & \textit{things} & \textit{stuff} & wall & floor & cab & bed & chair & sofa & tabl & door & wind & bkshf & pic & cntr & desk & curt & fridg & showr & toil & sink & bath & ofurn \\ \hline
      \multirow{3}{*}{\begin{tabular}[c]{@{}c@{}}PanopticFusion\\ w/o CRF\end{tabular}} & PQ & 29.7 & 26.7 & 56.7 & 37.5 & 76.0 & 18.6 & 29.1 & 37.8 & 38.2 & 29.5 & 13.8 & 14.1 & 13.0 & 26.5 & 8.3 & 14.9 & 11.6 & 38.0 & 28.8 & 72.4 & 33.3 & 28.0 & 24.3 \\

       & SQ & 71.2 & 71.4 & 69.5 & 62.3 & 76.7 & 69.4 & 68.5 & 69.3 & 72.3 & 70.1 & 74.6 & 69.9 & 70.7 & 72.9 & 65.0 & 60.6 & 70.5 & 75.3 & 75.8 & 79.2 & 71.9 & 74.0 & 75.3 \\
       & RQ & 41.1 & 36.8 & 79.6 & 60.2 & 99.0 & 26.8 & 42.5 & 54.6 & 52.8 & 42.1 & 18.5 & 20.1 & 18.4 & 36.3 & 12.8 & 24.6 & 16.4 & 50.4 & 37.9 & 91.3 & 46.4 & 37.8 & 32.2 \\ \hline
      \multirow{3}{*}{\begin{tabular}[c]{@{}c@{}}PanopticFusion\\ with CRF\end{tabular}} & PQ & 33.5 & 30.8 & 58.4 & 40.4 & 76.4 & 23.8 & 35.8 & 46.7 & 42.1 & 34.8 & 18.0 & 19.3 & 16.4 & 26.4 & 10.4 & 16.1 & 16.6 & 39.5 & 36.3 & 76.1 & 36.7 & 31.0 & 27.7 \\
       & SQ & 73.0 & 73.3 & 70.7 & 64.0 & 77.4 & 71.1 & 70.1 & 74.3 & 74.6 & 74.3 & 76.0 & 72.5 & 73.9 & 71.2 & 65.1 & 61.7 & 72.3 & 77.7 & 79.5 & 81.4 & 72.7 & 75.3 & 75.8 \\
       & RQ & 45.3 & 41.3 & 80.9 & 63.1 & 98.7 & 33.5 & 51.1 & 62.8 & 56.3 & 46.9 & 23.6 & 26.7 & 22.2 & 37.1 & 16.0 & 26.0 & 23.0 & 50.8 & 45.7 & 93.5 & 50.5 & 41.2 & 36.5 \\ \hline
      \end{tabular}
   }
\end{table*}


\begin{figure}[t]
   \centering
      \includegraphics[width=8.5cm]{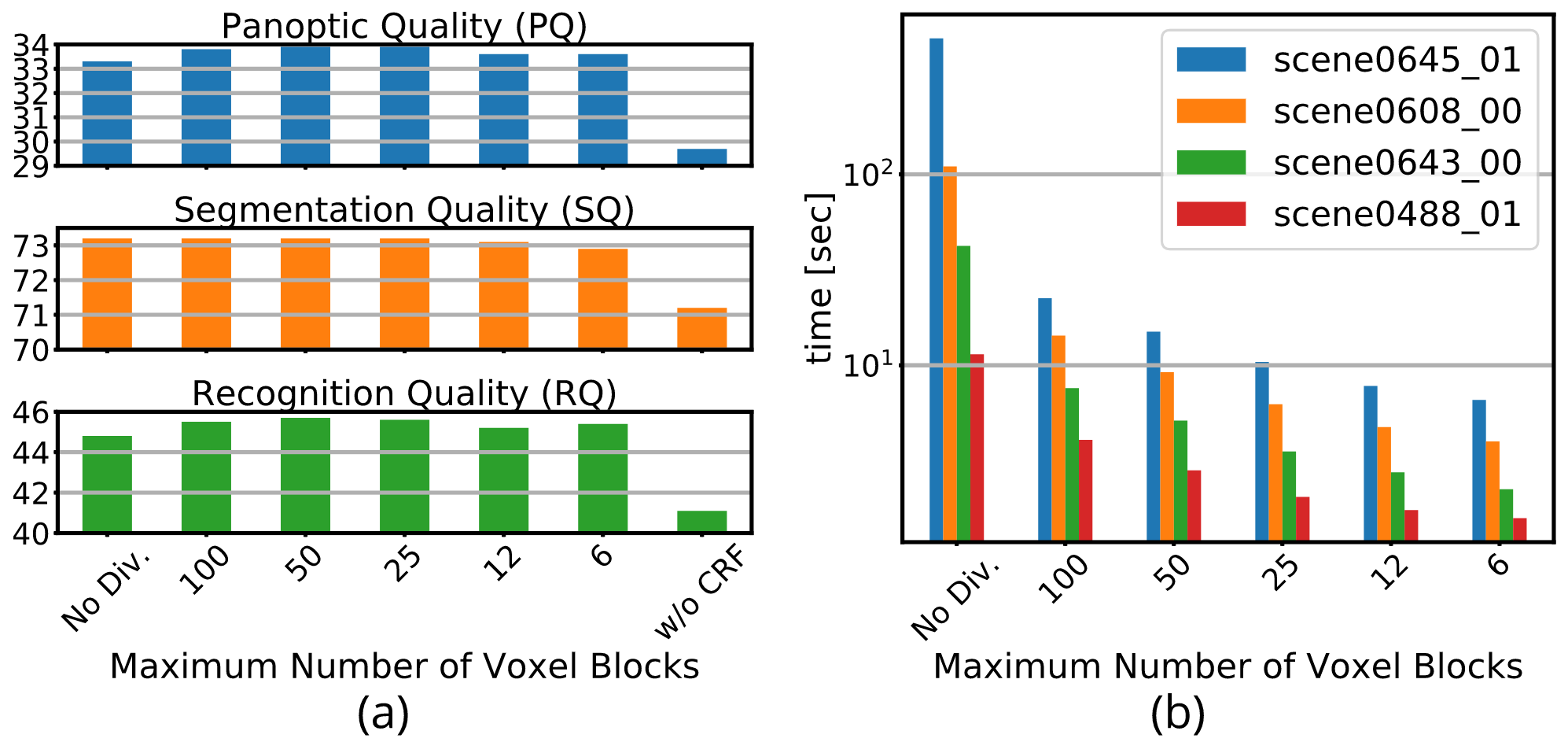}
      \caption{Results of the map regularization with the map division strategy. The relationship between the maximum number of voxel blocks and (a) recognition accuracy and (b) computational time. Note the computational time is shown in a logarithmic scale.}
      \label{fig_crf_with_map_division}
\end{figure}

\subsection{Evaluation of Map Regularization\label{sec_eval_crf}}
In this section, we evaluate the map regularization proposed in Section \ref{sec_map_regularization}.
First, we evaluated the effects of the map division on the recognition accuracy and computational time.
We used the open test set for the recognition accuracy and typical scenes in ScanNet v2 for the computational time.
The result is shown in Fig. \ref{fig_crf_with_map_division}.
Note that, in this experiment, we applied regularization to the pre-generated map as a post process to evaluate solely the effects of CRF.

As can been seen, the recognition performance was improved by the map regularization with the proposed unary potential approximation regardless of whether or not map division was used.
The results also show that the map division strategy drastically reduced the computational time without a decrease in recognition performance, compared with the case of building a CRF model for a whole map.

\begin{figure}[t]
   \centering
      \includegraphics[width=8.5cm]{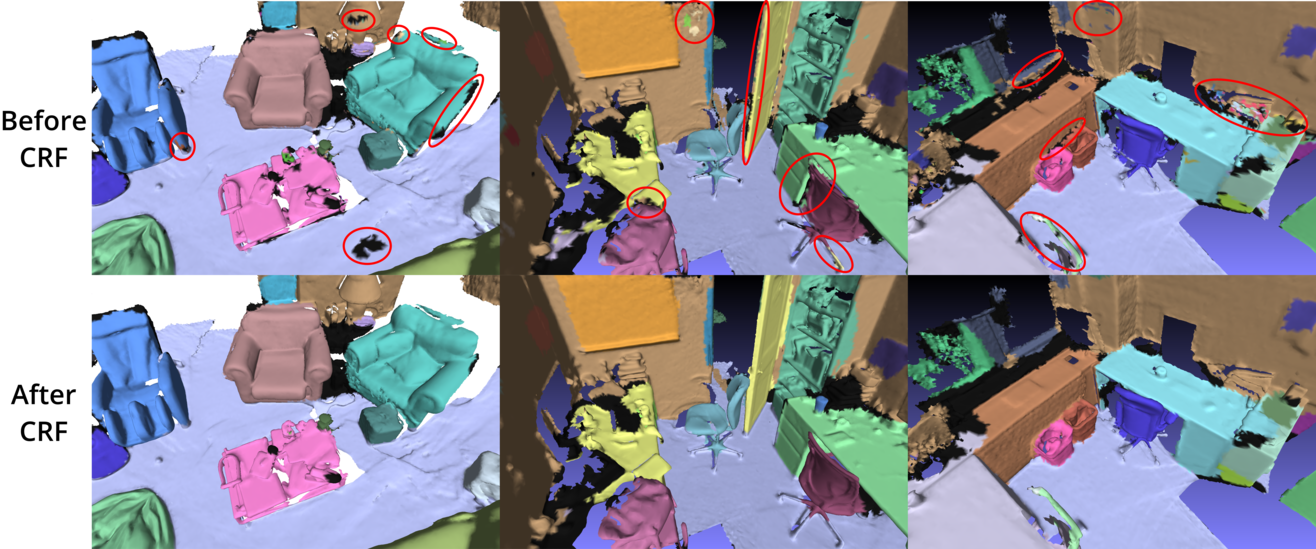}
      \caption{Qualitative results of map regularization. The noisy predictions within red circles are appropriately regularized, taking a spatial context into account.}
      \label{fig_crf_qualitative_reuslts}
\end{figure}

Based on the above results, our online system employed map regularization with the map division strategy.
We chose a maximum number of voxel blocks of 25 because of the better recognition accuracy and acceptable computational time.
Table \ref{table_panoptic_quality} shows the difference in recognition performance due to whether or not map regularization was used in online processing.
This result shows that the map regularization improved the recognition performance even when the system ran online.
Note that the scores of almost all the classes were boosted by the proposed regularization.
See Fig. \ref{fig_crf_qualitative_reuslts} for qualitative effects of the map regularization.

\subsection{Run-time Analysis}
Table \ref{table_runtime_analysis} shows computational times for each component of our system, which are measured on scene0645\_01, a typical large-scale scene in ScanNet v2 (shown in Fig. \ref{fig_panopticfusion}).
PSPNet and Mask R-CNN each run on GPUs, and the other components are processed on a CPU.
All components are basically processed in parallel. The throughput of our system is around 4.3 Hz, which is determined by Mask R-CNN, the bottleneck process of our system.
Although our current implementation is not highly optimized, our system is able to run at a rate allowing interaction.
Note that the computational time except for the map regularization does not depend on the scale of scenes nor the number of {\it things} because we utilize the raycasting approach for the integrations.
The processing time of the map regularization increases to about 10 seconds at the end of the sequence, but it could be reduced by processing only the voxel blocks near the camera frustum.

\begin{table}[t]
   \renewcommand{\baselinestretch}{0.8}
   \caption{Run-time analysis.}
   \label{table_runtime_analysis}
   \centering
   \scalebox{0.95}[0.95]{
      \begin{tabular}{llr}
         \hline
         \textbf{Frequency} & \textbf{Component} & \textbf{time} \\ \hline
         Every Mask R-CNN frames & PSPNet & 80 ms \\
          & Mask R-CNN & 235 ms \\
          & Panoptic label fusion & 2 ms \\ \cline{2-3} 
          & Reference panoptic label gen. & 19  ms \\
          & Panoptic label tracking & 9  ms \\
          & Volumetric integration & 139  ms \\
          & Probability integration & $\sim 1$  ms \\ \hline
         Every 10 sec. & Map regularization & 4.5 s \\ \hline
         Every 1 sec. & Mesh extraction & 14 ms \\ \hline \hline
         \textbf{Throughput} &  & 4.3 Hz \\ \hline
      \end{tabular}
   }
\end{table}

\section{APPLICATIONS}
\begin{figure}[t]
   \centering
      \includegraphics[width=8.5cm]{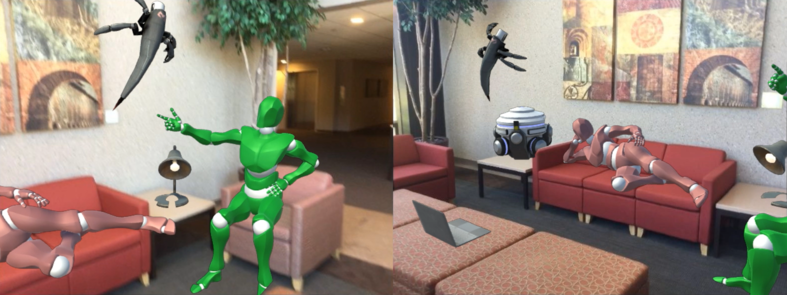}
      \caption{An example of AR application using a 3D panoptic map generated by PanopticFusion system.}
      \label{fig_ar_application}
\end{figure}

In this section, we demonstrate a promising AR application utilizing a 3D panoptic map generated online by the proposed system.
A 3D panoptic map reconstructed as 3D meshes allows us to realize the following visualizations according to the context of the scene:
\begin{itemize}
   \item Path planning on {\it stuff} regions such as floors and walls.
   \item Interaction with individual objects, or the {\it thing} regions.
   \item Interaction appropriate for the semantics of each region.
   \item Natural occlusion and collision visualization.
\end{itemize} 
We show an example of an AR application utilizing the above visualizations in Fig. \ref{fig_ar_application}. 
Humanoids and insect-type robots are able to locomote on the floor and wall meshes, respectively, according to the automatic path planning.
Additionally, the semantics of each object realizes context-aware interactions such that humanoids sit and lie on chairs and sofas, respectively, and CG objects appear on tables.
Moreover, we can naturally visualize the occlusion effects, which are important for AR, because the 3D meshes of the scene are extracted.
Note that, taking advantage of the accurately recognized 3D panoptic map, we can easily estimate the poses of seats of chairs and sofas, and top panels of tables by using simple normal- and curvature-based segmentation and plane detection.

We believe that our system is useful not only for AR scenarios but also for autonomous robots that explore scenes and manipulate objects.

\section{CONCLUSIONS}
In this paper, we have introduced a novel online volumetric semantic mapping system at the level of {\it stuff} and {\it things}.
It performs dense semantic labeling while discriminating individual objects, as well as large-scale 3D reconstruction and labeled mesh extraction thanks to the use of a spatially hashed volumetric map representation.
This was realized by pixel-wise panoptic label prediction and its volumetric integration with careful label tracking.
In addition, we constructed a fully connected CRF model with respect to panoptic labels and inferred it online with a novel unary potential approximation and a map division strategy, which further improved the recognition performance.
The experimental results showed that our system outperformed or compared well with state-of-the-art offline 3D DNN methods in terms of both 3D semantic and instance segmentation.
In future work, we plan to extend our system to ensure global consistency against long-term pose drift, to perform high-throughput mapping by network reduction, and to support dynamic environments.

We believe that the {\it stuff} and {\it things}-level semantic mapping will open the way to new applications of intelligent autonomous robotics and context-aware augmented reality that deeply interact with the real world.

\addtolength{\textheight}{-12cm}   






\bibliographystyle{plain}
\bibliography{paper}

\end{document}